\documentclass[runningheads,a4paper]{llncs}

\usepackage{amssymb}
\usepackage{graphicx}

\usepackage{url}
\urldef{\mailsa}\path|{alfred.hofmann, ursula.barth, ingrid.haas, frank.holzwarth,|
\urldef{\mailsb}\path|anna.kramer, leonie.kunz, christine.reiss, nicole.sator,|
\urldef{\mailsc}\path|erika.siebert-cole, peter.strasser, lncs}@springer.com|

\usepackage{booktabs} 
\usepackage{subfig}
\usepackage{multirow}
\usepackage{makecell}

\begin{document}

\mainmatter  

\title{Automatic Skin Lesion Segmentation Using Deep Fully Convolutional Networks
}

\titlerunning{Automatic skin lesion segmentation using deep fully convolutional networks}

%
%
\author{Hongming Xu%
\and Tae Hyun Hwang%
\thanks{Corresponding author email: hwangt@ccf.org}}
%

\institute{Department of Quantitative Health Sciences, Cleveland Clinic, Cleveland, OH, USA\\
}

%
%

\maketitle

\begin{abstract}
This paper summarizes our method and validation results for the ISIC Challenge 2018 - Skin Lesion Analysis Towards Melanoma Detection - Task 1: Lesion Segmentation
\end{abstract}

\section{Introduction}
Melanoma is the most aggressive type of skin cancer, which causes a majority of skin cancer deaths. The early diagnosis of melanoma can greatly reduce the mortality, as the melanoma in early stages can be cured with prompt excision~\cite{bi2017automatic,Xu2017}. Analysis of dermoscopy images plays an important role in the early detection of melanoma. However, human analysis is typically subjective, inaccurate and poorly reproducible even among experienced dermatologists, due to appearance variations of skin melanoma and similarities with other benign tumors (e.g., moles). Automatic analysis of dermoscopic images can assist dermatologists in clinical decision making, and even help patients to judge their skin lesions at home. To perform automatic analysis of dermascopic images, segmentation of skin lesions from surrounding normal regions is usually the first step. There have been a number of studies~\cite{Yu2017,Yuan2017,Zortea2017} that try to address skin lesion segmentation task, but this task is still a challenging problem due to image variations.

\section{Materials and Methods}

\subsection{Database description}

\begin{figure}[!htb]
    \centering
    \subfloat[]{{\includegraphics[width=2.6cm]{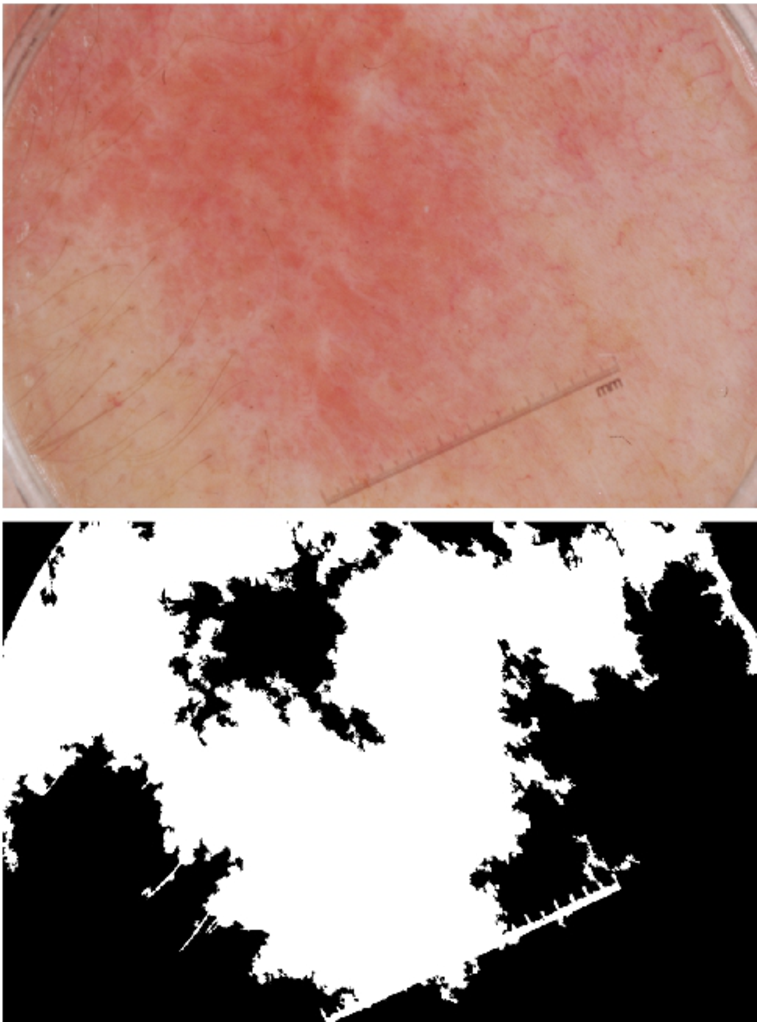} }}%
    \subfloat[]{{\includegraphics[width=2.6cm]{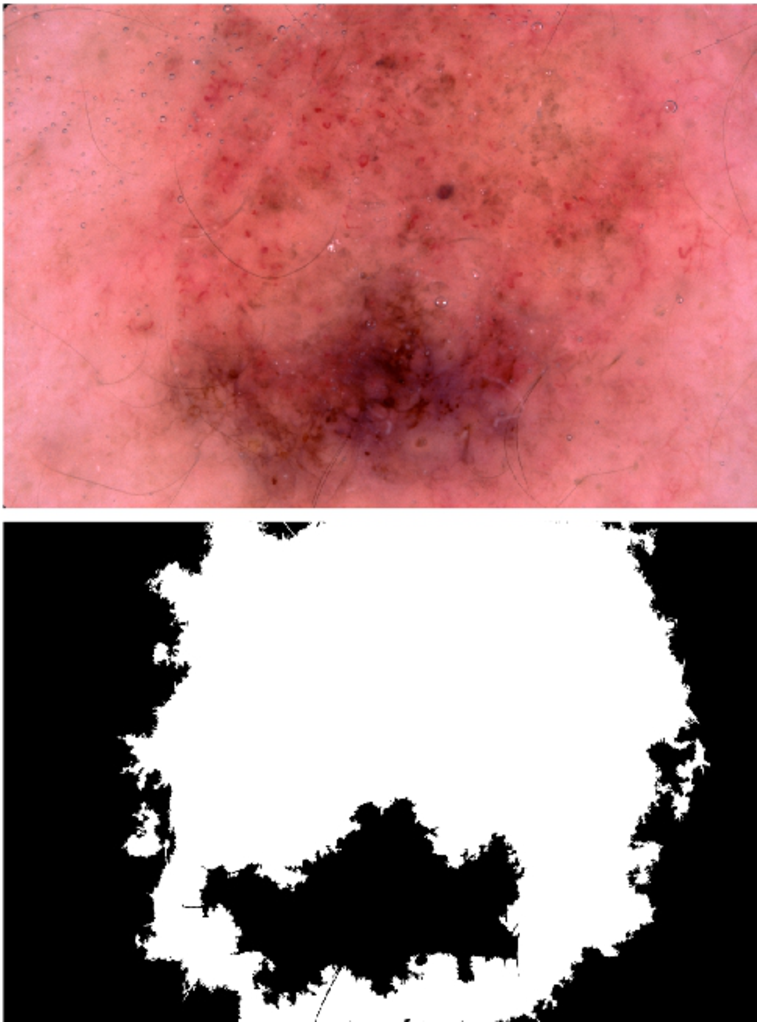} }}
    \subfloat[]{{\includegraphics[width=2.6cm]{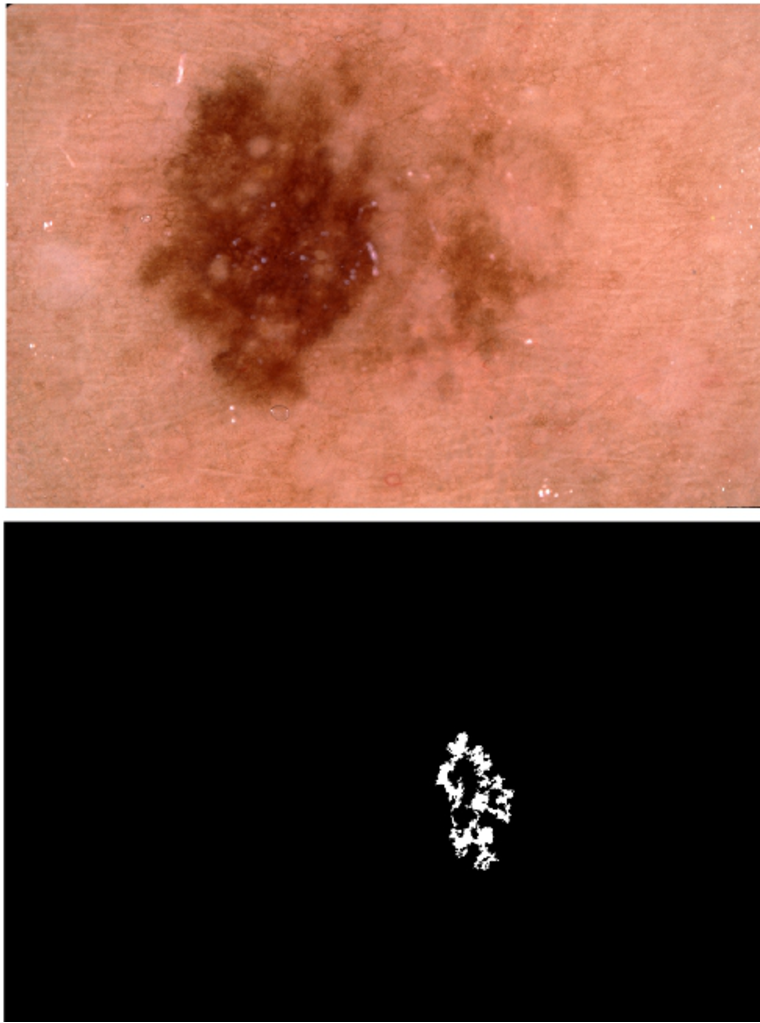} }}%
    \subfloat[]{{\includegraphics[width=2.6cm]{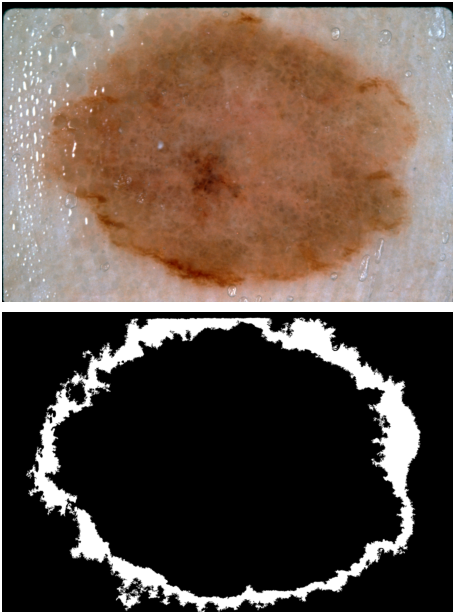} }}%
    \caption{Four excluded images with poor quality ground truths. (a) ISIC\_0012356. (b) ISIC\_0015013. (c) ISIC\_0015251. (d) ISIC\_0015353.}%
    \label{fig:example}
\end{figure}

All skin lesion dermoscopic images were download from ISIC 2018 challenge. We participated the Task 1: Lesion Segmentation, where training database originally includes 2594 skin lesion images in .jpg format, and corresponding 2594 ground truth segmentations in .png format. Based on visual examination, 4 images (see Fig.~\ref{fig:example}) are excluded from experiments due to poor quality ground truth segmentations (i.e., segmented skin lesion is not a continuous region). Therefore, there are 2590 skin lesion training images used in this study. These images have different sizes, ranging from hundreds by hundreds to thousands by thousands pixels. Besides training set, the challenge organizer provides a validation set with 100 images, and a testing set with 1000 images. The validation set is used for initial self-evaluation online, while the testing set is used for final performance rank of different research groups.
\graphicspath{{figs/}}

\subsection{Methods}

\subsubsection{Model architecture:} 
\begin{figure}[!htb]
    \centering
    \subfloat[]{{\includegraphics[width=0.6\textwidth]{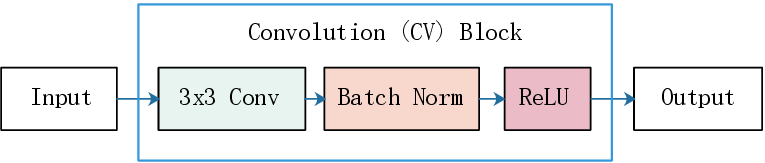} }}\
    \subfloat[]{{\includegraphics[width=0.67\textwidth]{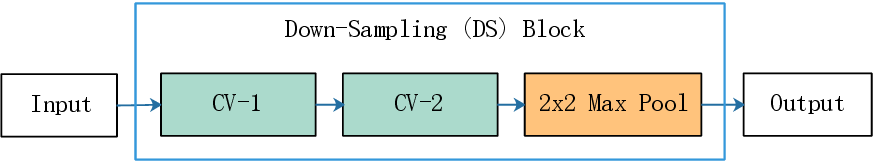} }}\
    \subfloat[]{{\includegraphics[width=0.79\textwidth]{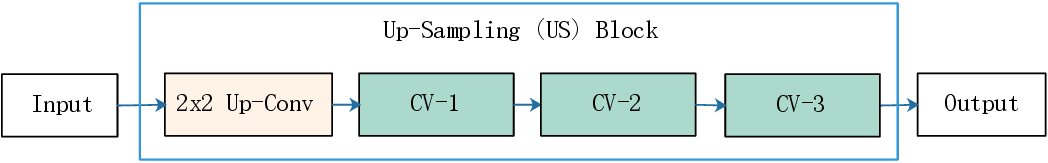} }}\
    \subfloat[]{{\includegraphics[width=0.88\textwidth]{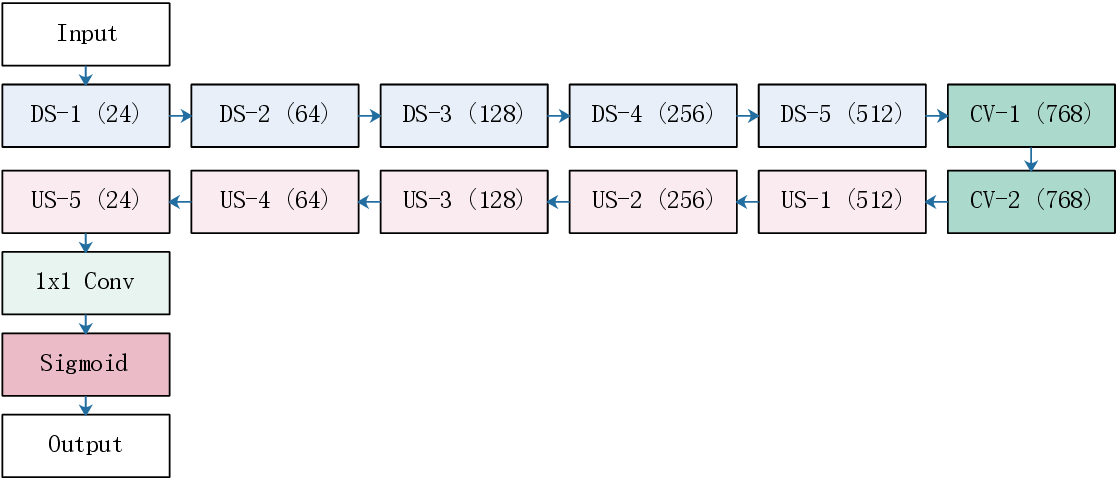} }}

    \caption{Illustration of model architecture. (a) Convolution (CV) block. (b) Down-sampling (DS) block. (c) Up-sampling (US) block. (d) Whole architecture. Note that in (d) numbers inside parenthesis are the numbers of feature channels.}%
    \label{fig:architexture}
\end{figure}

Motivated by the U-Net model~\cite{Olaf2015}, we train a deep fully convolutional network that maps the input dermoscopic image into a posterior probability map. The model architecture is shown in Fig.~\ref{fig:architexture}. Specifically, Fig.~\ref{fig:architexture}(a) shows the convolution (CV) block, which consists of 3x3 padded convolution, batch normalization and ReLU activation. Fig.~\ref{fig:architexture}(b) shows the down-sampling (DS) block, which consists of 2 CV blocks and 1 2x2 max-pooling layer. Fig.~\ref{fig:architexture}(c) shows the up-sampling (DS) block, which consists of 1 2x2 up-convolution layer and 3 CV blocks. Fig.~\ref{fig:architexture}(d) shows the whole architecture of our trained deep learning model, which includes 27 CV blocks, 5 up-convolution layers, 5 max-pooling layers and 1 output layer (1x1 convolution operation followed by sigmoid activation). Note that in Fig.~\ref{fig:architexture}(d) the numbers inside parenthesis are the number of feature channels. Overall, the whole network has about 28M trainable parameters.

\subsubsection{Training:}
Since most of images in the training set have a height-to-width ratio of 3:4, all training images are resized to 384x512 for input into our deep learning model. The network is trained using Adam optimization with a learning rate of ${10^{ - 4}}$ which is not changed throughout the training process. Mini-batches of size 8 are used. To improve the robustness of the trained model under a variety of image acquisition conditions, image augmentations are randomly performed on every mini-batch. The images are augmented by performing geometric transformations which include randomly rotation, scaling (up to 10\%) and shifting (up to 10\%). The training is allowed to perform 40 epoches, and each epoch is trained with 1295 steps. To reduce the training time, early stopping is performed, which stops training if there is no performance improvement after 5 epoches. To train the model, we design a loss function $L$ that incorporates binary cross entropy and Dice coefficient together, which is defined as below:

\begin{equation}
L =  - \sum\limits_{i,j} {\left[ {{g_{i,j}}\log \left( {{p_{i,j}}} \right) + \left( {1 - {g_{i,j}}} \right)\log \left( {1 - {p_{i,j}}} \right)} \right]}  + \left( {1 - \frac{{2 \times \sum\limits_{i,j} {{g_{i,j}}{p_{i,j}}} }}{{\sum\limits_{i,j} {{g_{i,j}}}  + \sum\limits_{i,j} {{p_{i,j}}} }}} \right)
\end{equation}
where ${{g_{i,j}}}$ and ${{p_{i,j}}}$ are the target and output for pixel $\left( {i,j} \right)$, respectively.

\subsubsection{Postprocessing} After obtaining the posterior map of validation image from trained model, we use a dual-thresholds method~\cite{Yuan2017} to generate a binary skin lesion mask. A relatively higher threshold (${t_h} = 0.8$) is first applied to select the most likely skin lesion region (i.e., with the largest average likelihood values). A lower threshold (${t_h} = 0.5$) is then applied on the posterior map to determine candidate skin lesion regions. The final skin lesion mask is determined as the candidate region that embraces the most likely skin lesion region (determined by the threshold $t_h$).

\section{Implementation and results}

Our method was implemented using Python and Keras library (with a Tensorflow backend). The experiments were conducted on a Alienware Area-51 R4 desktop with Intel (R) i7-7800X CPU 3.5GHz and 2 GPU of Nvidia GeForce GTX 1080 Ti with 11GB memory.

We trained 5 models using 2590 training images with 5 random training-to-validation split of 8:2. We tested the 5 models with 100 independent validation images. The obtained skin lesion masks of validation images are submitted to ISIC 2018 Challenge Website for online evaluations. Note that the online evaluation first compares each predicted segmentation with the corresponding ground truth segmentation using Jaccard index~\cite{Yuan2017}. The final segmentation score for each image is then assigned as Jaccard index value if it is above 0.65, otherwise the final score is assigned as 0. Table~\ref{result} lists the average segmentation scores for 100 validation images using 5 trained models. We run 5 models separately on 1000 testing images, and the obtained posterior maps from 5 models are averaged for postprocessing. The finally obtained 1000 skin lesion masks are submitted to ISIC 2018 Challenge Website for ranking.

\begin{table}[!htb]
  \caption{Performance on validation set using 5 trained models}
  \centering
  \label{result}
  \begin{tabular}{c|c|c|c|c|c}
    \Xhline{0.7pt}
    Models & Model-I & Model-II & Model-III & Model-IV & Model-IV\\
    \hline
    Segmentation Scores  & 0.738 & 0.731 & 0.724 & 0.730 & 0.749\\
  \Xhline{0.7pt}

\end{tabular}
\end{table}

\end{document}